\documentclass{article}
\usepackage{spconf,amsmath,graphicx}

\usepackage{CJKutf8}
\usepackage{booktabs}  % -----
\usepackage{float}     % -----
\usepackage{mathrsfs}
\usepackage{multirow}
\usepackage{makecell}
 \usepackage{hyperref}

\hypersetup{hidelinks,
	colorlinks=true,
	allcolors=black,
	pdfstartview=Fit,
	breaklinks=true}

\usepackage{color}

\title{Model Doctor for Diagnosing and Treating Segmentation Error
}
\name{Zhijie Jia$^{\dagger}$, Lin Chen$^{\dagger}$, Kaiwen Hu$^{\dagger}$, Lechao Cheng$^{\dagger\dagger}$\sthanks{Corresponding author.}, Zunlei Feng$^{\dagger}$, Mingli Song$^{\dagger}$}
\address{$^{\dagger}$ Zhejiang University\\ $^{\dagger\dagger }$ Zhejiang Lab}

\begin{document}
%\ninept
\maketitle
\begin{abstract}
\begin{CJK}{UTF8}{gbsn}
% 图像分割任务是一个重要的计算机视觉任务，在很多场景中得到了广泛应用，但目前的语义分割结果中仍存两类错误，语义类别错误和区域边界错误，即区域类别分类错误和物体区域完整分割错误，本文我们分别通过语义类别治疗和区域边界治疗方法，完成了对语义分割模型的治疗，进一步提升了语义分割模型在公开数据集上的分割能力。实验证明，我们提出的治疗方法可以针对多个分割模型在不同的数据集上获得有效的治疗效果。
\end{CJK}
Despite the remarkable progress in semantic segmentation tasks with the advancement of deep neural networks, existing U-shaped hierarchical typical segmentation networks still suffer from local misclassification of categories and inaccurate target boundaries. In an effort to alleviate this issue, we propose a Model Doctor for semantic segmentation problems. The Model Doctor is designed to diagnose the aforementioned problems in existing pre-trained models and treat them without introducing additional data, with the goal of refining the parameters to achieve better performance. Extensive experiments on several benchmark datasets demonstrate the effectiveness of our method. Code is available at \url{https://github.com/zhijiejia/SegDoctor}.
%Image segmentation task is an important computer vision task, which has been widely used in many scenarios, but there are still two types of errors in the current semantic segmentation results, semantic category error and region boundary error, that is, region category classification error and object region complete segmentation error. In this paper, we have completed the treatment of the semantic segmentation model through semantic category treatment and region boundary treatment methods respectively, and further improved the segmentation ability of the semantic segmentation model on the public dataset. Experiments show that our proposed treatment approach can achieve effective treatment results on different datasets for multiple segmentation models.
% \fzl{
% Image segmentation is a key task in the computer vision area with important applications such as scene understanding, video surveillance, medical image analysis, robotic perception, et al. Existing 
% }
\end{abstract}
\section{Introduction}
\label{sec:intro}
% 1. 分割很重要。
% 2. 难以debug
% 3. 已有方法xxx,但是无法做深度分割模型的debug
% 4. 如何来实现深度分割模型的debug? 通过实验分析发现：类别，边界错误。
% 5. in this paper, we xxxx
% 6. 实验结果显示xxx
% 7. contribution.
Image segmentation~\cite{voc2012, cityscapes, ade20k} is a crucial task in the computer vision field, with a wide range of applications~\cite{seg_survey1, seg_survey2}, including scene understanding, video surveillance, medical image analysis, robotic perception, and so on. 

% However, despite the focus on the design of deep convolutional neural networks in the current mainstream semantic segmentation technology, there is a neglect of the treatment and utilization of existing semantic segmentation models.
However, the current mainstream semantic segmentation techniques focus on the structural design of deep convolutional neural networks, but ignore the treatment and utilization of existing semantic segmentation models. In addition, the black box~\cite{understanding_black_box} structure of deep neural networks also contributes to the lack of ability to analyze problems from segmentation results, making it challenging to target errors and fine-tune the semantic segmentation model. 
\begin{figure}[htb]
\begin{minipage}[t]{1.0\linewidth}
  % \centering
  \centerline{\includegraphics[width=\linewidth]{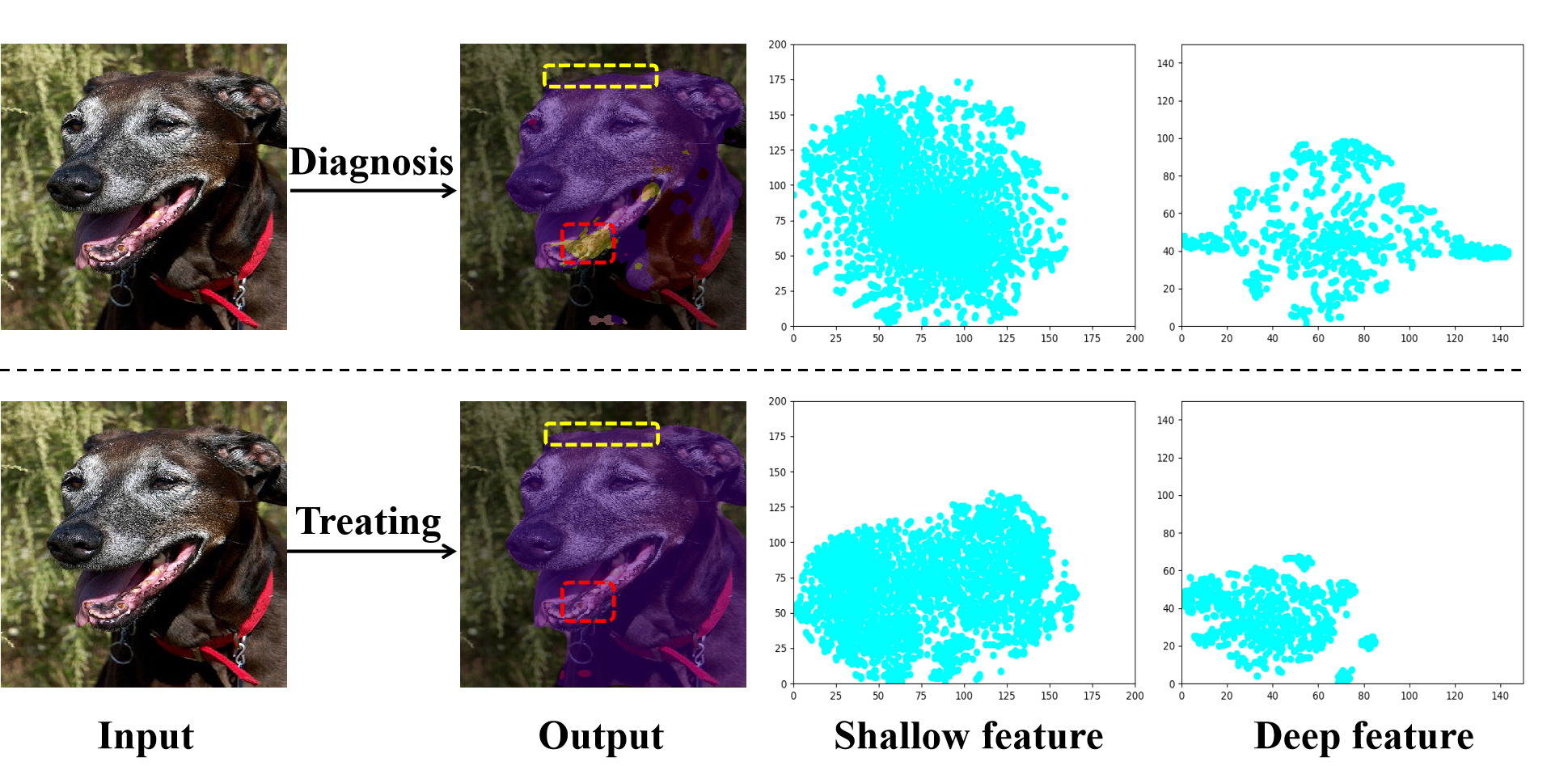}}
  % \vspace{-2.0em}
\end{minipage}
\caption{Feature analysis of semantic segmentation model. The content of the red box represents the category error, and the content of the yellow box represents the boundary error.}
\label{fig:feature_analysis}
\end{figure}      
There are currently model-interpretable methods that can assist in better understanding and analyzing models. However, much of the focus has been on visualizing model prediction results through techniques such as Class Activation Mapping (CAM)~\cite{cam}, Grad-CAM~\cite{grad-cam}, and Grad-CAM++~\cite{grad-cam++}. Through these methods, the patterns that the model prioritizes and the areas of input that the model pays more attention to can be identified. Additionally, some works utilize the interpretable random forests algorithm to dissect deep neural networks~\cite{lei2018understanding}, and decouple deep neural models, which facilitates rapid identification of the source and location of model errors. Nevertheless, these techniques cannot be applied directly and automatically to model treatment.

In the preliminary experiments, we find that errors in semantic segmentation models can generally be divided into two types: semantic category errors and regional boundary errors. Semantic category errors arise from the inclusion of feature errors in deep semantic features, resulting in category classification errors for certain regions. On the other hand, region boundary errors occur due to the lack of fine edge detail features in shallow texture features, resulting in lost boundary information.

In this paper, we introduce a Model Doctor to amend semantic category errors and regional boundary errors, respectively. As shown in Fig.~\ref{fig:feature_analysis}, we apply semantic category treatment to deep semantic features extracted by deep neural networks to bridge the gap within classes in deep features and force intra-class features to converge to the category center. For regional boundary treatment, we constrain shallow texture features at various levels to enhance internal feature constraints on objects and preserve more edge detail features. Exhaustive experiments demonstrate that incorporating the proposed method with several semantic segmentation models leads to improved performance on commonly used datasets. Our contributions can be summarized as follows:
\begin{itemize}
    \item We present a Model Doctor for diagnostic treatment segmentation models, which can be plugged into existing convolutional segmentation models.
    \item Semantic category treating strategy and region boundary treating strategy are designed to address semantic category errors and region boundary errors, respectively.
    \item Extensive experiments showcase that the proposed semantic segmentation model treating method can effectively boost the performance of existing semantic segmentation models.
\end{itemize}

\section{Related Work}
\label{sec:related work}
Due to the complexity and ambiguity of deep neural networks, humans cannot give exact explanations for their behavior. At present, the interpretability methods of deep models are mainly divided into two categories~\cite{interpretability_survey}: Post-hoc interpretability analysis method and Ad-hoc interpretable modeling method. Post-hoc interpretability analysis method is an interpretable analysis of deep models that have been trained; Ad-hoc interpretable modeling method mainly builds deep models into interpretable models to ensure that the inferences of the models are interpretable. Post-hoc interpretability analysis method mainly include seven categories of techniques, such as feature analysis~\cite{inverting_visual_representations, visualizing_higher_layer_features, convergent_learning}, model checking~\cite{understanding_black_box, towards_transparent_systems}, salient expression~\cite{Greedy_function_approximation, peeking_black_box}, surrogate modeling~\cite{decision_tree}, advanced mathematics analysis~\cite{beyond_finite_layer}, case interpretation~\cite{case_based_reasoning}, and text interpretation~\cite{visual_alignments}. Ad-hoc interpretable modeling method mainly includes two types of methods: interpretable representation~\cite{infogan} and model improvement~\cite{exact_consistent}. However, the above methods mainly focus on model interpretation and cannot achieve automatic diagnosis and optimization of model defects. Recently, Feng et al.~\cite{modeldoctor} proposed model doctor for the optimization of classified convolutional neural networks, but due to the difference between the segmentation model and the classification model architecture, this method cannot be applied to the semantic segmentation model.

\section{Method}
\label{sec:method}
In this paper, we present a novel model therapeutic approach for semantic segmentation models, designed to address the inadequacies in semantic category classification and boundary refinement of these models.

%（诊断）
\subsection{Segmentation Error Diagnosis}
In the preliminary experiments, we find that semantic segmentation models are prone to regional boundary problems and category classification problems, and different model problems are related to different feature errors.% in the model.
\subsubsection{Semantic category error}
The semantic segmentation model is typically composed of an encoder and a decoder, where the encoder is responsible for extracting image features and the decoder is responsible for restoring image edge details. Given an input image $I$, the output feature map of the last layer of the encoder is $M^e$, computed as $M^e = Encoder(I)$, where the shape of $M^e$ is $(N, C, H, W)$, where $N$ is bach size, $C$ is the number of channel and $(H, W)$ is the feature map size, and the vectors of each $(1, C, 1, 1)$ in $M^e$ correspond to a patch in the original image. The deep features extracted by the encoder $M^e$, possess a wealth of deep semantic information and semantic category information. The widening gap between the deep feature vectors signifies that the semantic category information of the corresponding patches is no longer equivalent, leading to subsequent classification errors.

\subsubsection{Regional boundary error}
\label{secsecsec:boundary_error}
The extracted image features $\{ M^e_{1}, M^e_{2}, M^e_{3}, ..., M^e_{l}, ..., M^e_{L}\}$ of the encoder exhibit distinct attributes at various depths, where $L$ is the maximum layer number in the encoder. While shallow image features $M^e_{l}$ are rich in edge detail information, they lack semantic intricacies; Conversely, deep image features $M^e_{l}$ are abundant in semantic information but deficient in edge detail. The extensive semantic features of $M^e_{l}$ enable the model to perform efficient class classification, whereas the edge details present in $M^e_{l}$ aid in partial reconstruction of the object's edge details by the decoder. 

Hence, during the decoding phase, the shallow and deep feature maps $\{M^e_{l}\}^L_{l=1}$ are concatenated and processed by a convolutional function $\mathcal{F}_{conv}$, to produce the feature map $M^d_{i+1}$ of $i$-th layer as follows:
\begin{equation}\label{eq:decoder_output}
M^d_{i} = \mathbf{Concat}(M^e_{l},  M^d_{i-1}), l \in \{1,2,3, ..., L\},
    %F_{i+1} = \mathbf{Concat}(M^e_{l},  M^d_{i+1}), M^d_{i+1} = \mathcal{F}_{conv}(F_{i}),
\end{equation}
%\begin{equation}\label{eq:decoder_output}
 %   Output_{i} = \mathcal{F}_{conv}(F_{i}),
%\end{equation}
where the initial input feature map $M^d_{1}$ of encoder is the output feature map $M^e_{L}$ of the encoder.
However, if the shallow feature $M^d_{i}$ of the decoder contains errors in shallow detail information, the model will miss crucial detail information during the upsampling process, rendering it insensitive to the object's edge area and incapable of producing fine-grained edge details of the object.
\subsection{Segmentation Error Treatment}
%Based xxx, 我们分别设计了xxx、xxx，来实现 分类、边界错误的治疗。
In light of the aforementioned observations, we have developed a segmentation error diagnosis and treating method that encompasses both semantic category correction and regional boundary rectification, aiming at addressing the classification and boundary errors of the semantic segmentation model.
\begin{figure*}[!htb]
\begin{minipage}[t]{\linewidth}
  \centerline{\includegraphics[width=\linewidth]{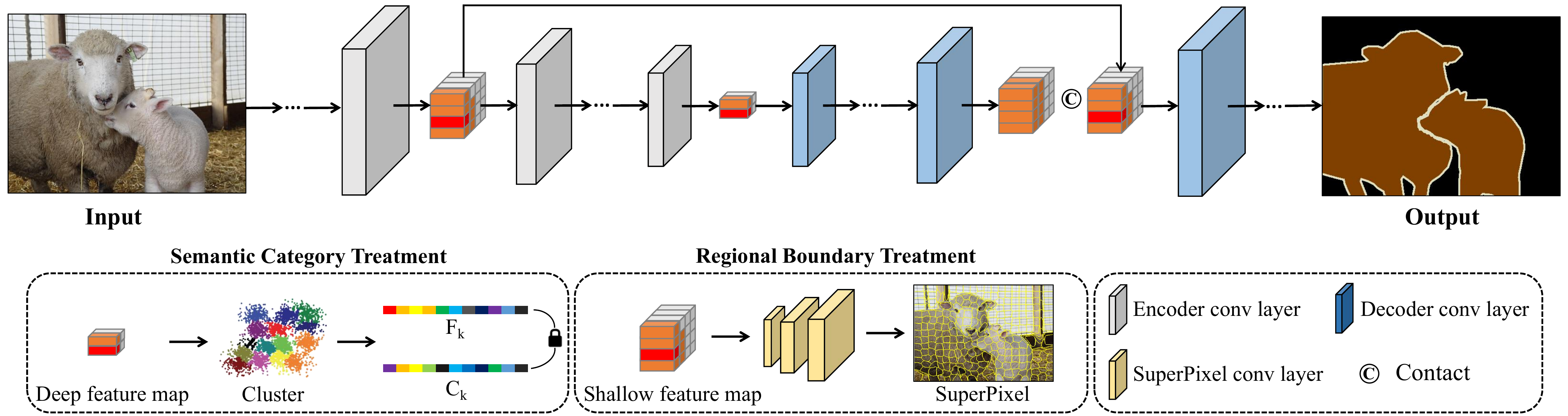}}
\end{minipage}
\caption{
The framework of the proposed method, which is comprised of two parts: the semantic category treatment applied to the deep features, and the regional boundary treatment applied to the shallow features.
}
\label{fig:framework_fig}
% \vspace{-5mm}
\end{figure*}

\subsubsection{Treating Category error}
Consequently, in order to mitigate the impact of semantic category errors on deep features, we devise a category constraint technique for treating semantic category error. It constrains the deep features of the model by minimizing inter-class variations and maximizing intra-class similarity. 
To achieve this, the cluster center $C_k$ for the $k$-th class in 
cluster $D_k$ is computed to represent the central tendency of features within each class and provides a basis for comparison with other feature vectors. The cluster center $C_k$ is calculated as follows:
\begin{equation}\label{eq:kmeans_loss}
    % min\sum_{x \in D_{k}}||x - C_{k}||^{2}
    \mathop{\arg\min}\limits_{C_{k}}\sum_{R_k \in D_{k}}||R_k - C_{k}||^{2},
\end{equation}
where $R_k$ is the feature representation in cluster $D_k$.

In the context of deep features, the image feature for a given class $k$ is denoted as $R_{k}$. To alleviate semantic errors and improve the model's classification accuracy, a feature distance constraint is imposed to force the intra-class image features to gravitate towards the centroid of the class cluster $C_k$, which can mitigate intra-class feature divergence. The feature distance penalty $\zeta_{sim}$ is calculated as follows:
\begin{equation}\label{eq:sim_loss}
    \zeta_{sim} = 1 - \mathcal{D}(C_{k}, R_{k}), \mathcal{D}(C_{k}, R_{k}) = \frac{C_{k} \cdot R_{k}}{||C_{k}|| \times ||R_{k}||},
\end{equation}
%\begin{equation}\label{eq:sim_loss}
  %  \mathcal{D}(C_{k}, F_{k}) = \frac{\vec{C_{k}} \cdot \vec{F_{k}}}%{||C_{k}|| \times ||F_{k}||},
%\end{equation}
where `$\cdot$' denotes vector multiplication, $\mathcal{D}(C_{k}, R_{k})$ represents the feature distance between the feature representation $R_{k}$ and the cluster center $C_{k}$.

\subsubsection{Treating boundary error}
In accordance with the information presented in Section~\ref{secsecsec:boundary_error}, if the shallow image features contain erroneous texture feature information, this can result in inaccuracies in the decoder's fine edge reconstruction. To address this, superpixel technology is incorporated as superpixel branch, which is a coarse segmentation method that helps preserve edge details and enforce consistency within shallow image features. The SpixelFCN algorithm proposed in~\cite{superpixel} is a noteworthy implementation of superpixel segmentation that leverages a fully convolutional network to achieve rapid and remarkable results. In this work, we drew inspiration from SpixelFCN to devise the superpixel branch, aiming to preserve the shallow texture features. The superpixel branch is assembled by a block consisting of three conv-bn-relu layers, which performs the upsampling operation and generates the link probability connecting the pixel to the neighboring superpixels.

For shallow feature map $M^e_{l}$, the superpixel branch $\mathcal{F}_{sup}$ predicts the probability of $p$ being associated with surrounding superpixels as follows:
\begin{equation}\label{eq:spixel_branch1}
    p = \sigma(\mathcal{F}_{sup}(M^e_{l})),
\end{equation}
where $\sigma(\cdot)$ represents the sigmoid function.
Then the reconstruction of pixel feature $\mathbf{f'(\cdot)}$ and pixel coordinates $\mathbf{v'}$ are calculated as follows:
\begin{equation}\label{eq:spixel_branch2}
    \mathbf{v'} = \sum_{s \in \mathcal{N}_{\mathbf{v}}}\frac{\sum_{\mathbf{v}:s \in \mathcal{N}_{\mathbf{v}}} \mathbf{v} \cdot p}{\sum_{\mathbf{v}:s \in \mathcal{N}_{\mathbf{v}}}p} \cdot p,
\end{equation}
\begin{equation}\label{eq:spixel_branch3}
    \mathbf{f'(v)} = \sum_{s \in \mathcal{N}_{\mathbf{v}}}\frac{\sum_{\mathbf{v}:s \in \mathcal{N}_{\mathbf{v}}}\mathbf{f(v)} \cdot p}{\sum_{\mathbf{v}:s \in \mathcal{N}_{\mathbf{v}}}p} \cdot p,
\end{equation}
where  $\mathbf{v} = [x, y]^{T}$ denotes the original  pixel's position, and $\mathcal{N}_{\mathbf{v}}$ is the set of surrounding superpixels of $\mathbf{v}$. The penalty function of the superpixel branch is divided into two parts: feature constraint and coordinate constraint, which is specified as follows:
\begin{equation}\label{eq:loss_spixel}
    \zeta_{sp} = \sum_{\mathbf{v}}CE(\mathbf{f(v)}, \mathbf{f'(v)}) + \frac{m}{s}||\mathbf{v} - \mathbf{v'}||_{2},
\end{equation}
where $\mathbf{f}(\cdot)$ represents one-hot encoding vector of semantic label. $s$ denotes the superpixel sampling interval, and $m$ is a weight-balancing term, and $CE(\cdot, \cdot)$ denotes Cross-Entropy. 

Overall, with the shallow feature map $M^e_{l}$ as input, the first terms of $\zeta_{sp}$ encourages the trained superpixel branch $\mathcal{F}_{sup}$ to group pixels with similar category property, and the second term enforces the superpixels to be spatially compact.

\subsection{Overview}
\label{sssec:framework}
Finally, the total loss function adopted is the original Cross-Entropy loss $\zeta_{ce}$ combined with the category error loss $\zeta_{sim}$ and the boundary error loss $\zeta_{sp}$ as follows:
\begin{equation}\label{eq:total_loss}
Loss = \zeta_{ce} + \alpha  \zeta_{sim} + \beta  \zeta_{sp},
\end{equation}
where $\alpha$ and $\beta$ denote the balance parameters.

%The results, as depicted in Table~\ref{tab:comparing_method}, 

\begin{table}[!t]
    \centering
    \caption{The performance on different models and datasets.}
    \resizebox{\linewidth}{!}{
    \begin{tabular}{c|c|c|c|c}
            \toprule
            {Dataset $\rightarrow$} & \multicolumn{2}{c}{{VOC 2012}} & \multicolumn{2}{|c}{{Cityscapes}}\\
            \midrule
            {Method $\downarrow$} &{Origin}&{+Treatment} &{Origin}&{+Treatment} \\
            \midrule
            {FPN}              &{61.7} &62.5(\textbf{+0.8}) &{66.5} &{67.9}(\textbf{+1.4}) \\
            {UNet}            &{54.0} &55.2(\textbf{+1.2}) &{69.5} &{70.1}(\textbf{+0.6})  \\
            {CCNet}          &{57.1} &58.7(\textbf{+1.6}) &70.8 &72.0(\textbf{+1.2})  \\
            {PSPNet}        &{68.1} &69.0(\textbf{+0.9}) &72.8 &74.1(\textbf{+1.3})  \\
            {Deeplab v3+}    &{67.3} &68.4(\textbf{+1.1}) &74.2 &74.9(\textbf{+0.7})  \\
            \bottomrule
    \end{tabular}
    }
    \label{tab:comparing_method}
\end{table}

\begin{table}[!t]
    \centering
    \caption{The ablation study on different treating strategies.}
    \begin{tabular}{cc}
        \toprule
        \makecell[l]{\textbf{Method}}       &\makecell[r]{\textbf{mIoU}}\\
        \cmidrule(r){1-2}
        \makecell[l]{UNet}                  &\makecell[r]{54.0}\\
        \makecell[l]{+ Treating category}   &\makecell[r]{54.4 \textbf{(+0.4)}}\\
        \makecell[l]{+ Treating boundary}   &\makecell[r]{54.8 \textbf{(+0.7)}}\\
        \makecell[l]{+ Treating category \& boundary}                 &\makecell[r]{55.2 \textbf{(+1.2)}}\\
        \bottomrule
    \end{tabular}
    \label{tab:ablation_study}
\end{table}

\section{Experiment}
\label{sec:experiment}
\subsection{Dataset and Experiment setting}
% \begin{CJK}{UTF8}{gbsn}
% 1.数据集 (VOC、Cityscape)、实验设置 \\
% \end{CJK}
\textbf{Dataset.} Our experimental evaluation is performed on two publicly available datasets, namely the PASCAL VOC 2012~\cite{voc2012} dataset and Cityscapes dataset~\cite{cityscapes}. The PASCAL VOC 2012 dataset, a semantic segmentation dataset with 20 categories, comprises 10,582 images in its training set and 1,449 images in its validation set. The Cityscapes dataset, a driving dataset for panoramic segmentation with 19 categories, comprises 2,975 images in the training set and 500 images in the validation set. \\
\noindent \textbf{Experiment Setting.} During the training of models, we randomly crop images to 512 $\times$ 512 (VOC) and 512 $\times$ 1024 (Cityscapes) and utilize horizontal and vertical flipping augmentations. The batch size is set to 8 for all datasets, and the optimization is performed using Stochastic Gradient Descent (SGD). The initial learning rate is set at 0.01 and the cosine annealing rate decay policy is employed. The balance parameters are set as follows: $\alpha = 1$ and $\beta = 0.01$. 
The performance of the semantic segmentation is reported using the mean Intersection over Union (mIoU) metric.

\subsection{Compatibility with Existing Segmentation Models}
In the experiment, we adopt some mainstream segmentation network to verify the effectiveness of the proposed method. Results in Table~\ref{tab:comparing_method}
demonstrate that the proposed approach is able to enhance the performance of different models on the PASCAL VOC 2012 dataset and the Cityscapes dataset.

\begin{figure}[!t]
\centering
\begin{minipage}[b]{0.9\linewidth}
  \centering
  \centerline{\includegraphics[width=\linewidth]{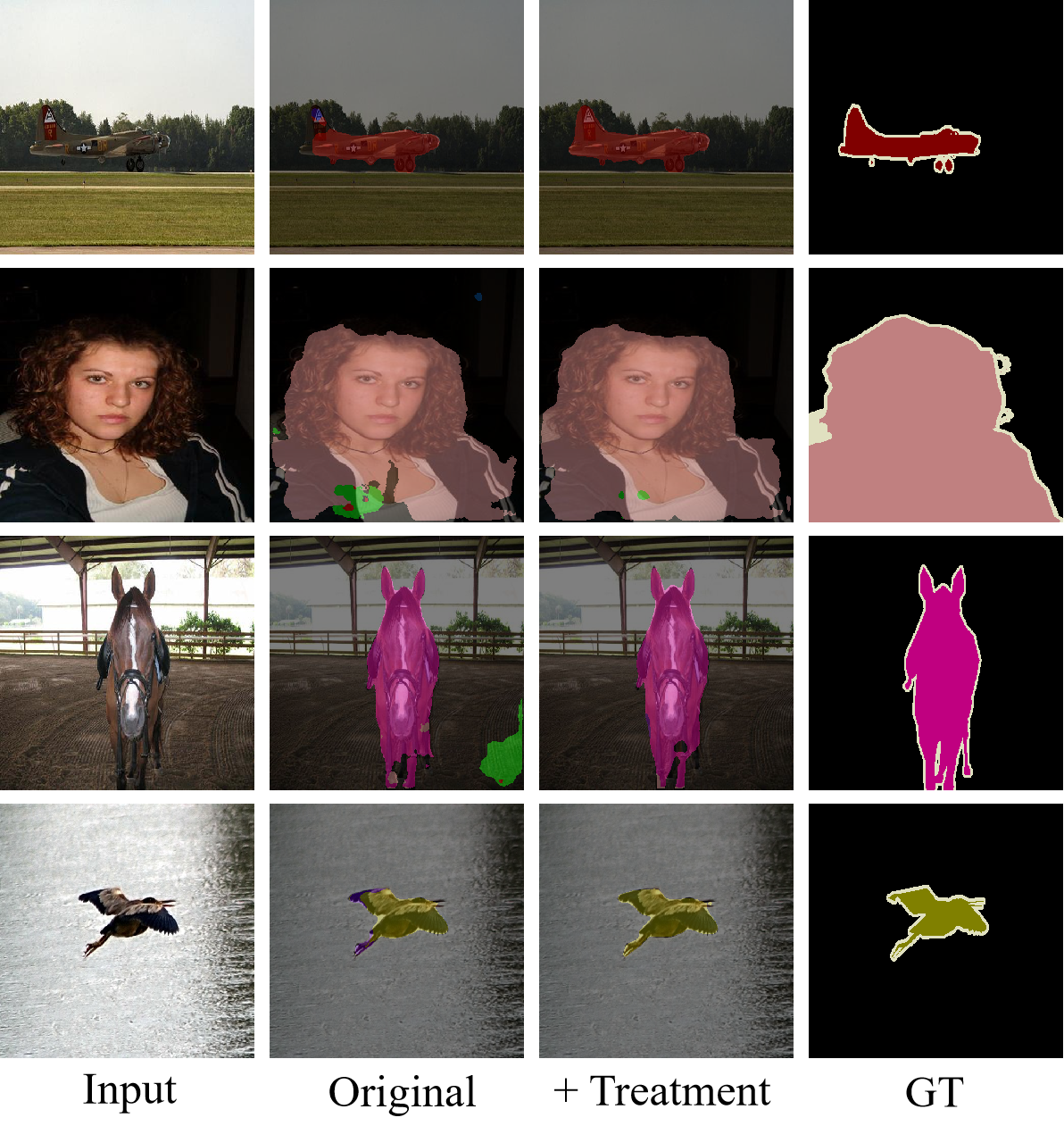}}
\end{minipage}
\caption{Visual results on PASCAL VOC 2012 dataset.}
\label{fig:visual_result}
\vspace{-5mm}
\end{figure}

\subsection{Visual Results}
We demonstrate the efficacy of the proposed method by incorporating it into the UNet network on the VOC 2012 dataset, resulting in improved semantic segmentation performance. As depicted in Fig.~\ref{fig:visual_result}, our method produces more accurate and nuanced structures, as evidenced by several visualizations from the VOC 2012 validation set.
\subsection{Ablation Study}
In this section, we conduct the ablation study on two treatment strategies. The ablation study experiment is conducted with UNet on the VOC 2012 dataset. As shown in Table~\ref{tab:ablation_study}, the semantic category treatment strategy and regional boundary treatment strategy both effectively enhance the performance of the segmentation model.

\section{Conclusion}
In this paper, a new method called Model Docter is introduced to address semantic category errors and regional boundary errors in semantic segmentation. Semantic category treatment is applied to deep semantic features extracted by deep neural networks to reduce gaps within classes and correct misclassifications. Regional boundary treatment is imposed on shallow texture features to enhance internal feature constraints and preserve edge detail features. The proposed approach has been tested on several datasets and models and can be combined with other models for further refinement.
\label{sec:conclusion}

% \section{COPYRIGHT FORMS}
% \label{sec:copyright}

% You must submit your fully completed, signed IEEE electronic copyright release
% form when you submit your paper. We {\bf must} have this form before your paper
% can be published in the proceedings.

\vfill\pagebreak

\bibliographystyle{IEEEbib}
\bibliography{ref.bib}

\end{document}